\documentclass[journal]{IEEEtran}
\usepackage{amsmath,amsfonts}
\usepackage{algorithmic}
\usepackage{algorithm}
\usepackage{array}
\usepackage[caption=false,font=normalsize,labelfont=sf,textfont=sf]{subfig}
\usepackage{textcomp}
\usepackage{stfloats}
\usepackage{url}
\usepackage{verbatim}
\usepackage{graphicx}
\usepackage{cite}
\usepackage{booktabs}
\usepackage{tabularx}
\usepackage{multirow}

\usepackage[table]{xcolor}
\usepackage{arydshln}
\usepackage{amssymb}
\usepackage[hidelinks]{hyperref}
\definecolor{baselinebg}{RGB}{255,255,255}
\definecolor{oursbg}{RGB}{226,232,255}

\definecolor{gaincolor}{RGB}{0,150,85}
\definecolor{losscolor}{RGB}{192,57,43}

\newcommand{\scoreplain}[1]{\makebox[\linewidth][c]{#1}}
\newcommand{\scoregain}[2]{\makebox[\linewidth][c]{\bfseries#1\rlap{\hspace{.08em}\raisebox{.15ex}{\scriptsize\textcolor{gaincolor}{+#2$\uparrow$}}}}}

\newcolumntype{Z}[1]{%
  >{\hsize=#1\hsize
    \linewidth=\hsize
    \centering\arraybackslash}X%
}
\newcolumntype{Y}{>{\centering\arraybackslash}X}
\newcommand{\dg}[1]{\textcolor{gaincolor}{#1}}

\begin{document}
\bstctlcite{BSTcontrol}
\title{ReUnit: Multi-Granularity Visual Unitization for Long Video Understanding}

\author{
  Biao Tang,
  Xu Chen,
  Shuxiang Gou,
  Jingyi Yuan,
  Yuhan Zhang,
  Deyu Meng,~\IEEEmembership{Member,~IEEE},
  and Chenqiang Gao,~\IEEEmembership{Member,~IEEE}
\thanks{Biao Tang, Jingyi Yuan and Chenqiang
Gao are with the School of Intelligent Systems Engineering, Shenzhen
Campus of Sun Yat-sen University, Shenzhen, Guangdong 518107, P.R. China.
Corresponding author: Chenqiang Gao (e-mail: gaochq6@mail.sysu.edu.cn).}

\thanks{Xu Chen is with the School of Computer Science and Technology, Harbin Institute of Technology, Shenzhen, 518055, China (email: lanncx@gmail.com)}


\thanks{Shuxiang Gou is with the Shenzhen Institute for Advanced Study, University of Electronic Science and Technology of China, Shenzhen 518110, China (email: shuxianggou@std.uestc.edu.cn).}

\thanks{Yuhan Zhang is with the School of Flexible Electronics, Sun Yat-sen University, Shenzhen, Guangdong, China.}

\thanks{Deyu Meng is with the School of Mathematics and Statistics, Xi'an Jiaotong University, Xi'an, Shaanxi 710049, P.R. China.}


}

\markboth{}%
{Tang \MakeLowercase{\textit{et al.}}: ReUnit: Multi-Granularity Visual Unitization for Long Video Understanding}


\maketitle

\begin{abstract}

Long-video understanding is constrained by the limited visual input capacity of video multimodal large language models (Video-MLLMs).
Existing methods mainly optimize which content to retain, while the
presentation of retained content often remains fixed. As a result, the same
balance between spatial detail and content coverage is imposed across the
entire visual input.
We propose ReUnit, a training-free and query-aware framework that jointly
determines which content to retain and how it should be presented.
Guided by frame-level query relevance, ReUnit constructs and allocates visual
units at multiple presentation granularities. More relevant content receives
finer presentation, while broader context is represented more compactly.
It realizes these granularities using visual units that carry one, four, or
nine source frames and are each rendered as a standard image.
Across four benchmarks and visual input budgets from 4 to 32 units, ReUnit achieves the highest average score at every tested budget.
It improves over Uniform Sampling by 8.3--10.1 points on average, and the gains transfer across three additional Video-MLLM families. Project resources are available at
\url{https://github.com/charon525/ReUnit}.

\end{abstract}


\section{Introduction}

\IEEEPARstart{V}{ideo} multimodal large language models (Video-MLLMs) have recently made notable progress on complex multimodal reasoning tasks such as video question answering~\cite{yu2024prompting,zhao2026videoexpert,tang2026videounderstanding}. However, as the input grows from short clips to long videos that last tens of minutes or even hours, the limited context capacity and increasing inference cost quickly become major bottlenecks~\cite{you2024toward,Tang2025aks,ma2026gift}. A long video usually contains thousands of frames. Dense encoding produces far more visual tokens than the model can accommodate
and substantially increases inference cost. Simply reducing the sampling
density, however, may miss the key content needed to answer the query~\cite{wang2025divico}. Extending the context window~\cite{shu2025videoxl,chen2025longvila}, compressing visual tokens~\cite{wang2025divico}, or summarizing the video into text~\cite{ma2025drvideo,you2024toward} can ease this pressure, but they may introduce additional cost, require model modifications, or sacrifice visual detail. Constructing the visual input instead leaves the downstream model unchanged
and can be used with different Video-MLLMs. How to organize a long video into a compact and effective visual input sequence under a limited visual input budget has therefore become a central challenge in long video understanding.

\begin{figure}[t]
    \centering
    \includegraphics[width=\linewidth]{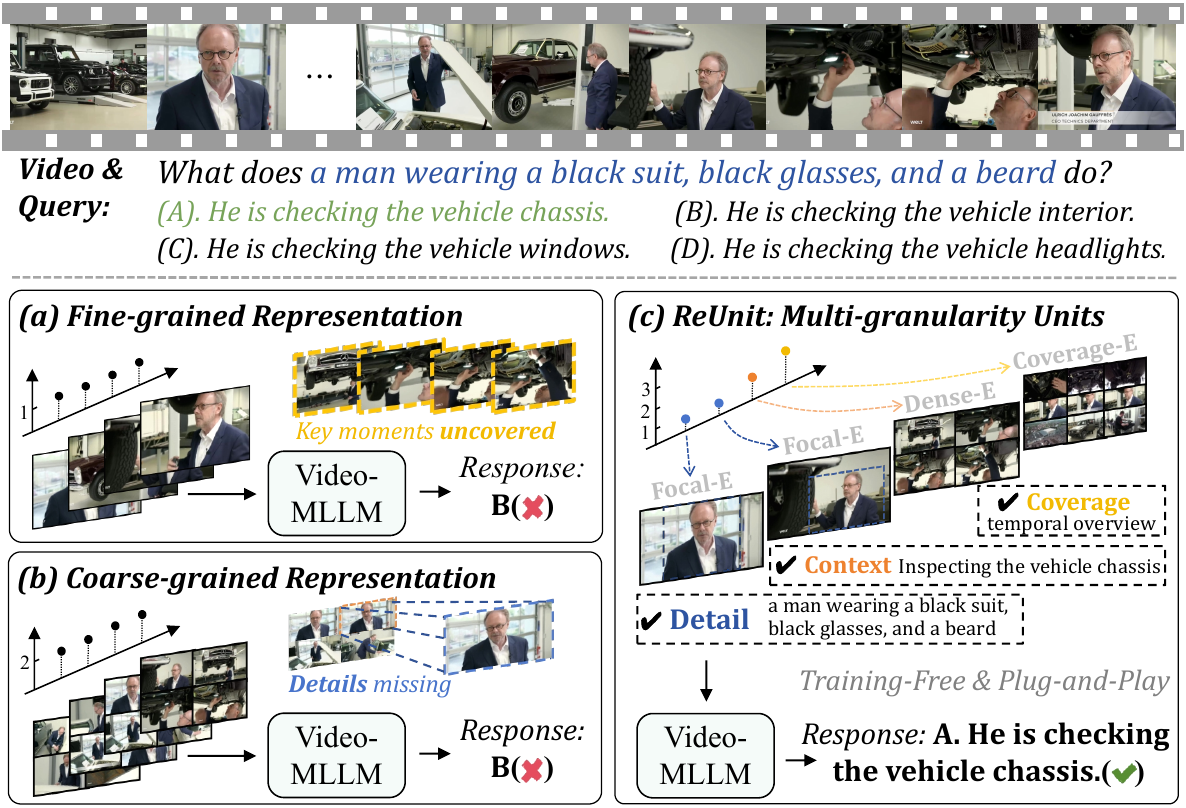}
    \caption{A comparison of visual-input construction strategies under a limited visual-input budget, with the query ``What does a man wearing a black suit, black glasses, and a beard do?'' \textbf{(a)} The fine-grained representation has a high per-frame cost and covers limited content, so it misses the chassis-inspection moment and answers incorrectly. \textbf{(b)} The coarse-grained representation trades uniform multi-frame compression for coverage and loses key details, so it also answers incorrectly. \textbf{(c)} ReUnit combines three visual-unit types, Focal-E, Dense-E, and Coverage-E, according to content importance, thereby preserving fine detail while extending content coverage and producing the correct answer.}
    \label{fig:intro}
\end{figure}

For this challenge, most existing methods can be grouped into two strategies. 
The first selects key frames or segments using query relevance, content irreplaceability, semantic boundaries, and other cues~\cite{sun2025mdp3,Tang2025aks,ma2026gift,chen2026wavelet,kim2026request,peng2026qca,liu2025bolt,tan2026thinkclipsample,tang2025tspo,yao2025gens}. Despite different selection criteria and budget allocation strategies, the retained content is typically presented as complete visual inputs, inherently limiting the amount of visual information covered by the input sequence under a fixed budget (Fig.~\ref{fig:intro}(a)).
The second strategy compresses several candidate frames into one visual input~\cite{doorenbos2026videopanels} to reduce the average representation cost per frame. However, applying the same compression to all content forces key and secondary content to share the same representation capacity, which can obscure fine-grained details (Fig.~\ref{fig:intro}(b)).
Both strategies therefore rely on a single presentation granularity for the entire visual input: fine granularity preserves detail but limits coverage, whereas coarse granularity expands coverage at the risk of detail loss. Under a limited visual input budget, this imposes the same balance between spatial detail and content coverage on all retained content.

To move beyond the limitation of a single granularity, we need to reconsider how retained content should be presented. The key observation is that video content contributes unequally to a given query, yet a single-granularity representation treats all retained content alike. In film narrative, presentation detail varies with narrative importance. Key moments are shown in close-up to reveal critical details. Less central content can be presented in more compact forms: split screens allow multiple scenes to be presented within a single frame. This differentiation lets both fine detail and the overall narrative thread fit within limited screen time. The same principle motivates long-video input construction: content of different importance need not share the same presentation granularity. The problem is therefore not only \textit{which content to retain}, but also \textit{how the retained content is presented}: can presentation granularity vary with content importance instead of remaining uniform?

To answer this question, we propose \textbf{ReUnit}, a training-free and
query-aware framework in which query-conditioned content importance guides
both which content is retained and the granularity at which it is presented.
ReUnit achieves different presentation granularities by varying the number of
source frames grouped into each visual unit (Fig.~\ref{fig:intro}(c)).
Concretely, it defines three visual-unit types that each consume one unit of
the visual input budget but operate at different granularities. The
\textbf{Focal-Evidence Unit (Focal-E)} devotes a full visual input to a single
frame, preserving fine detail. The \textbf{Dense-Evidence Unit (Dense-E)} uses a
$2\times2$ spatiotemporal unitization to carry multi-moment information from a
local segment. The \textbf{Coverage-Evidence Unit (Coverage-E)} uses a
$3\times3$ spatiotemporal unitization to provide an overview over a larger
temporal span. Guided by query relevance, ReUnit assigns these units over
non-overlapping intervals: Focal-E to the most relevant content, Dense-E to
unoccupied local segments, and Coverage-E across the remaining temporal
regions. Ordered by their timestamps, these units form a temporally ordered
visual input sequence that concentrates fine detail on highly query-relevant
content while using coarser units to represent broader temporal context within
the same visual input budget.


Our main contributions are as follows:
  \begin{itemize}
      \item We formulate long-video input construction under a fixed visual input budget as query-guided visual unitization, jointly determining \textit{which content to retain} and \textit{how the retained content is presented}.
      \item We propose ReUnit, a training-free and query-aware framework that
      constructs three visual-unit types with different presentation granularities:
      Focal-E, Dense-E, and Coverage-E. Each consumes one unit of the visual input
      budget. A priority-based span-allocation procedure assigns these units over
      non-overlapping temporal intervals and orders them into a temporally ordered
      visual input sequence (Section~\ref{sec:method}).
      \item We evaluate ReUnit on four long-video understanding benchmarks, MLVU~\cite{zhou2025mlvu}, LongVideoBench~\cite{wu2024longvideobench}, LVBench~\cite{wang2025lvbench}, and VideoMME~\cite{fu2025videomme}, under four visual input budgets (Section~\ref{sec:exp}). ReUnit achieves the highest average score across these benchmarks at every tested budget, and its gains transfer across three additional Video-MLLM families.
  \end{itemize}

\section{Related Work}

\subsection{Long Video Understanding with Video-MLLMs}
Video-MLLMs have achieved strong performance on video question answering and related multimodal reasoning tasks~\cite{cheng2024videollama2,lin2024videollava,yu2024prompting,wu2024collaborative,zhao2026videoexpert,gong2025spatial,tang2026videounderstanding}. However, extending Video-MLLMs  to long videos remains challenging because longer videos generate substantially more visual tokens, straining the available context capacity and increasing inference cost. Existing approaches address this limitation either by increasing the effective context available to the model or by  reducing the visual representation cost of the input.

One direction extends the effective context available to long-video inputs through long-context transfer or context expansion~\cite{zhang2024longva,shu2025videoxl,chen2025longvila,yan2026internvideo3,li2026videochat3}. Another reduces the visual representation cost through memory-based processing and visual-token compression, including compact representations and token reduction~\cite{song2024moviechat,he2024malmm,shen2025longvu,li2024llamavid,liu2024stllm,wang2025divico}. A related line converts long videos into textual or hierarchical semantic representations, allowing subsequent reasoning to operate on condensed descriptions rather than dense visual inputs~\cite{you2024toward,ma2025drvideo,wang2025videotree}. These approaches can improve the tractability of long-video processing, but they generally introduce additional components for context management, visual compression, or semantic summarization around the downstream model.

A complementary direction reorganizes the visual input before it enters the model. SlowFast-LLaVA uses slow and fast visual streams with different sampling rates and spatial resolutions, whereas Video Panels arranges ordered frames into panel
images~\cite{xu2024slowfastllava,doorenbos2026videopanels}. Both designs modify the visual input before it is processed by the downstream Video-MLLM. Within this broader class of visual-input designs, the following section considers methods that use the query to determine which visual content is presented to the model.

\subsection{Query-Aware Visual Input Construction}

A large body of work constructs compact visual inputs by selecting frames or segments that are useful for a given query. Some methods learn query-conditioned sampling policies or selectors from additional supervision~
\cite{yao2025gens,tang2025tspo,kim2026request}. Training-free methods more commonly use pretrained vision-language models to rank content according to query relevance, coverage, or content irreplaceability~\cite{liu2025bolt,Tang2025aks,ma2026gift}. Other methods account for the temporal or content structure of the video when selecting frames or allocating the visual input budget across temporal regions~\cite{sun2025mdp3,chen2026wavelet,peng2026qca,ziruiz2026focus,tan2026thinkclipsample}. These methods refine the selection of content under a limited visual input budget. However, the selected content is typically presented as individual frames in a common representation format. The selection decision therefore primarily determines which content to retain, rather than how the retained content is presented.

Some recent methods combine query-aware selection with different spatial
resolutions for individual frames~\cite{Zhang2025qframe,wu2026vqos,lu2026gaussian}.
Such methods allocate different amounts of spatial detail across selected
frames while retaining individual frames as the representation unit. They may
also depend on whether the downstream model supports variable-resolution
inputs. ReUnit instead constructs units at different presentation
granularities by varying the number of source frames grouped into each unit.
Query relevance then guides their allocation. In this way, ReUnit jointly
determines \textit{which content to retain} and \textit{how the retained
content is presented}.

\begin{figure*}[t]
    \centering
    \includegraphics[width=\textwidth]{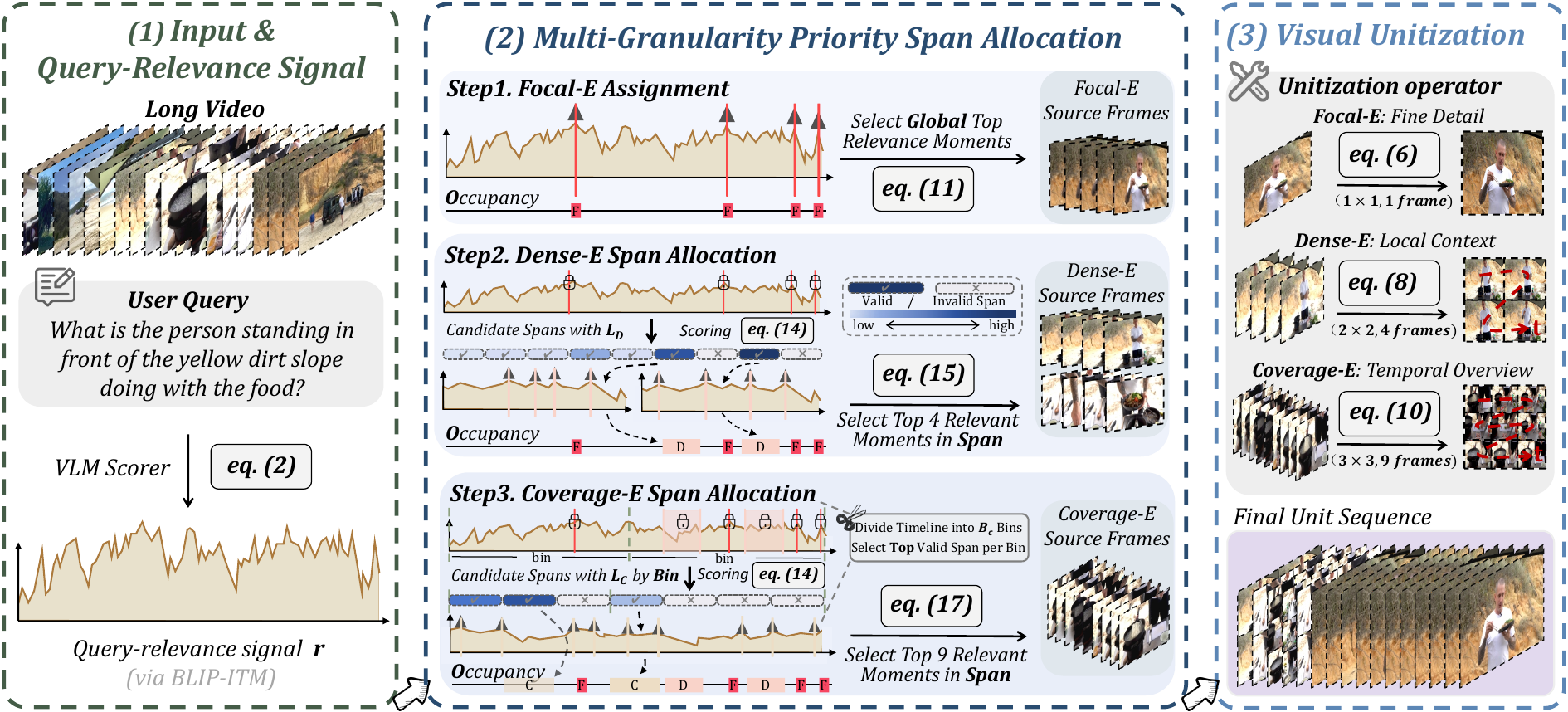}
    \caption{Overview of ReUnit. Guided by a frame-level query-relevance signal, ReUnit reorganizes a long video into a sequence of visual units that each consume one unit of the visual input budget but operate at different granularities: Focal-E (1 frame), Dense-E ($2{\times}2$, 4 frames), and Coverage-E ($3{\times}3$, 9 frames). Subject to a temporal occupancy constraint, the priority-based allocation process first selects the most relevant frames as Focal-E source frames, then allocates unoccupied local spans to Dense-E, and finally distributes Coverage-E spans across the remaining temporal regions. The selected source frames are then unitized at their corresponding granularities. The resulting units are ordered by timestamp and provided to a downstream Video-MLLM.}
    \label{fig:method}
\end{figure*}

\section{Method}
\label{sec:method}
Rather than introducing a new query-relevance scoring mechanism, ReUnit takes query-relevance scores for sampled frames as given. Under a fixed visual input budget, it uses these scores to determine both the retained content and the presentation granularity of each visual unit. As shown in Fig.~\ref{fig:method}, the allocation process selects the most relevant frames for Focal-E, allocates unoccupied local spans to Dense-E, and selects Coverage-E spans from the remaining unoccupied temporal regions. These decisions define the type and source frame indices of each unit specification. The resulting unit specifications are ordered by their representative timestamps. Visual unitization then converts the ordered specifications into image inputs for the downstream Video-MLLM. The next three subsections formulate the input-construction problem, define the three visual units and the unitization process, and describe the default allocation procedure.

\subsection{Problem Formulation}
\label{subsec:problem}

Given a long video $V$, a text query $q$, and a visual input budget $B$,
ReUnit constructs a temporally ordered sequence of $B$ unit specifications:
\begin{equation}
\mathcal{U}=\{u_i\}_{i=1}^{B},\qquad
u_i=(z_i,\mathcal{I}_i,\tau_i).
\end{equation}
Here, $B$ denotes the number of visual inputs available to the downstream
Video-MLLM. $z_i$ denotes the unit type, $\mathcal{I}_i$ contains the source
frame indices carried by the unit, and $\tau_i$ denotes its representative
timestamp. Visual unitization converts each specification $u_i$ into an image
input $\tilde{u}_i$, yielding the visual input sequence
$\widetilde{\mathcal{U}}=\{\tilde{u}_i\}_{i=1}^{B}$, which is provided together
with the query $q$ to the downstream Video-MLLM.

To obtain the query-relevance signal, ReUnit first samples $V$ at a fixed
frame rate to obtain $\mathcal{X}=\{x_t\}_{t=1}^{N}$, where $N$ is the number
of sampled frames. A frozen vision-language matching model then computes the
relevance between each sampled frame and the query:
\begin{equation}
r_t=s(x_t,q),\qquad t=1,\ldots,N,
\qquad
\mathbf{r}=[r_1,\ldots,r_N].
\end{equation}
The resulting frame-level query-relevance signal $\mathbf{r}$ guides both
source-frame selection and the presentation granularity assigned to each unit.

\subsection{Visual Unitization}
\label{subsec:unitization}

Visual unitization converts each unit specification $u_i$ into one image input $\tilde{u}_i$. The unit type $z_i\in\{F,D,C\}$ corresponds to Focal-E, Dense-E, or Coverage-E. The set $\mathcal{I}_i$ contains the source frame indices carried by the unit. The representative timestamp of $u_i$ is
\begin{equation}
\tau_i=\frac{1}{|\mathcal{I}_i|}\sum_{t\in\mathcal{I}_i}t.
\end{equation}
Each specification is converted into one image input at the same resolution. Focal-E preserves the detail of a single moment.Dense-E represents several moments from a local span. Coverage-E provides a broader temporal overview.

ReUnit defines a unitization granularity $g_z$ for each unit type $z$,
with $g_F=1$, $g_D=2$, and $g_C=3$. The corresponding number of source
frames is $m_z=g_z^2$. For unitization, the indices in $\mathcal{I}_i$ are ordered by time:
\begin{equation}
\mathcal{I}_i=\{t_1,t_2,\ldots,t_{m_{z_i}}\},\qquad t_1<t_2<\cdots<t_{m_{z_i}}.
\end{equation}
The spatiotemporal unitization operator then produces the image input
\begin{equation}
\tilde{u}_i=\rho_{g_{z_i}}(x_{t_1},x_{t_2},\ldots,x_{t_{m_{z_i}}}).
\end{equation}
When $g_{z_i}=1$, the operator produces an image from a single source frame. When $g_{z_i}>1$, the operator places the source frames row by row within one image according to their temporal order. All unit types are rendered at a common resolution that satisfies the input-resolution constraint of the visual encoder. For Dense-E and Coverage-E, this common resolution additionally ensures that the internal boundaries of the $2\times2$ and $3\times3$ layouts align with the patch grid of the visual encoder. The three unit types therefore each count as one visual input under the budget, although they carry different numbers of source frames.

\subsubsection{Focal-Evidence Unit}
Focal-E corresponds to the unitization granularity $g_F=1$ and carries one source frame. For a selected frame $x_t$, its specification and image input are
\begin{equation}
u_F(t)=(F,\{t\},t),\qquad \tilde{u}_F(t)=\rho_1(x_t).
\end{equation}
Focal-E preserves the spatial detail of a single frame without combining it with other frames. During allocation, ReUnit assigns Focal-E to the moments most relevant to the query.

\subsubsection{Dense-Evidence and Coverage-Evidence Units}

Under a fixed visual input budget, a sequence of one-frame visual units can cover only a limited portion of a long video. Dense-E and Coverage-E address this limitation by arranging multiple source frames from a selected span $I=[s,e)$ into one image. Each resulting image counts as one visual input under the budget.

Given a selected span $I$, Dense-E carries four source frames from the span:
\begin{equation}
\begin{aligned}
&\mathcal{I}_D(I)=\{t_1,t_2,t_3,t_4\},\\
&t_1<t_2<t_3<t_4,\qquad t_j\in I.
\end{aligned}
\end{equation}
Its specification and image input are
\begin{equation}
\begin{aligned}
u_D(I)&=(D,\mathcal{I}_D(I),\tau_D(I)),\\
\tilde{u}_D(I)&=\rho_2(x_{t_1},x_{t_2},x_{t_3},x_{t_4}).
\end{aligned}
\end{equation}
Here, $\tau_D(I)$ follows the representative timestamp defined above. Dense-E represents four moments from a local span within one visual input.

Given a selected span $I$, Coverage-E carries nine source frames from the span:
\begin{equation}
\begin{aligned}
&\mathcal{I}_C(I)=\{t_1,t_2,\ldots,t_9\},\\
&t_j\in I, \quad t_1<t_2<\cdots<t_9.
\end{aligned}
\end{equation}
Its specification and image input are
\begin{equation}
\begin{aligned}
u_C(I)&=(C,\mathcal{I}_C(I),\tau_C(I)), \\
\tilde{u}_C(I)&=\rho_3(x_{t_1},x_{t_2},\ldots,x_{t_9}).
\end{aligned}
\end{equation}
Here, $\tau_C(I)$ follows the same representative timestamp definition. Coverage-E carries nine source frames within one visual input and is designed to provide broader temporal coverage than Dense-E.

The source frames in Dense-E and Coverage-E need not be consecutive. They are selected from the same span, so their temporal separation is bounded by the span length. Each unit therefore represents several moments from one part of the video rather than frames drawn from across the whole timeline. Section~\ref{subsec:allocation} describes how ReUnit selects these spans and source frames using query relevance and temporal occupancy.

\subsection{Multi-Granularity Priority Span Allocation}
\label{subsec:allocation}

After obtaining the query-relevance signal $\mathbf{r}$, ReUnit divides the budget $B$ among the three unit types and selects the source frames carried by each unit. It assigns Focal-E first so that the most relevant moments receive the finest presentation granularity. Dense-E is then allocated to unoccupied local spans. Coverage-E is allocated last by selecting spans from the remaining unoccupied temporal regions. The overall procedure is summarized in Algorithm~\ref{alg:reunit}.

Given a total visual input budget $B$, ReUnit divides the budget according
to a predefined allocation ratio, yielding integer unit counts
$B_F$, $B_D$, and $B_C$ for Focal-E, Dense-E, and Coverage-E, respectively,
with $B_F+B_D+B_C=B$. ReUnit maintains an occupancy set $\mathcal{O}$ containing the sampled frame
indices covered by previous assignments. A subsequent frame or span is
available only if its indices do not intersect $\mathcal{O}$. Here, a span
$I=[s,e)$ also denotes the set of sampled frame indices it covers. When a
span is assigned to Dense-E or Coverage-E, the entire span is added to
$\mathcal{O}$.

\textbf{Focal-E assignment.}
ReUnit assigns the $B_F$ Focal-E units one at a time. At each step, it selects the unoccupied sampled frame with the highest query relevance:
\begin{equation}
t^\star=
\operatorname*{arg\,max}_{t\in\{1,\ldots,N\}\setminus\mathcal{O}} r_t.
\end{equation}
The selected index defines the Focal-E specification
\begin{equation}
u_F(t^\star)=(F,\{t^\star\},t^\star).
\end{equation}
ReUnit then adds $t^\star$ to $\mathcal{O}$. This assigns the finest presentation granularity to the moments with the highest relevance.

After Focal-E assignment, a candidate span is available only when it does not intersect $\mathcal{O}$. Dense-E and Coverage-E are therefore allocated only to unoccupied temporal regions.

\textbf{Candidate spans.}
For $z\in\{D,C\}$ with $B_z>0$, a candidate span must contain at least $m_z=g_z^2$ sampled frames. The span is also bounded so that the source frames of one unit remain within a local temporal region. ReUnit sets the span length to
\begin{equation}
L_z=\max\left(m_z,\min\left(\kappa m_z,\left\lfloor\frac{N}{B_z}\right\rfloor\right)\right).
\end{equation}
The term $\kappa m_z$ bounds the range available to one unit. The term $\lfloor N/B_z\rfloor$ adjusts the span length according to the video length and the allocated budget. The outer maximum guarantees that a span contains enough sampled frames to construct one unit.

ReUnit constructs candidate spans $\mathcal{C}_z$ of length $L_z$ with
stride $L_z$ along the sampled sequence. If a shorter range remains at the
end of the video, ReUnit adds one additional span ending at the final sampled
frame. Only candidate spans that do not intersect $\mathcal{O}$ are considered during allocation. The source frames of each Dense-E or Coverage-E unit therefore come from one bounded temporal span.

\begin{algorithm}[t]
\caption{Multi-Granularity Priority Span Allocation}
\label{alg:reunit}
\begin{algorithmic}[1]
\REQUIRE Relevance signal $\mathbf{r}$, budget $B$, allocation ratio
$(\alpha_F,\alpha_D,\alpha_C)$, span expansion factor $\kappa$
\ENSURE Temporally ordered unit specifications $\mathcal{U}$

\STATE Convert $(\alpha_F,\alpha_D,\alpha_C)$ to integer counts
$(B_F,B_D,B_C)$, ensuring $B_F+B_D+B_C=B$ and at least one unit per type;
assign any rounding remainder to $B_C$
\STATE Initialize $\mathcal{U}\leftarrow\emptyset$ and
$\mathcal{O}\leftarrow\emptyset$

\STATE \textbf{Stage 1: Focal-E assignment}
\FOR{$i=1$ \TO $B_F$}
    \STATE Select the highest-relevance unoccupied frame $t^\star$
    \STATE Add $u_F(t^\star)$ to $\mathcal{U}$ and $t^\star$ to $\mathcal{O}$
\ENDFOR

\STATE \textbf{Stage 2: Dense-E span allocation}
\STATE Compute $L_D$ and construct candidate spans $\mathcal{C}_D$
\FOR{up to $B_D$ available spans}
    \STATE Select the highest-scoring valid span $I^\star$
    \STATE Select the $m_D$ most relevant frames in $I^\star$
    \STATE Add the resulting Dense-E unit to $\mathcal{U}$ and
    $I^\star$ to $\mathcal{O}$
\ENDFOR

\STATE \textbf{Stage 3: Coverage-E span allocation}
\STATE Compute $L_C$ and construct candidate spans $\mathcal{C}_C$
\STATE Divide the timeline into $B_C$ bins and group spans by midpoint
\FOR{each temporal bin $G_j$ with an available valid span}
    \STATE Select the highest-scoring valid span $I_j^\star$
    \STATE Select the $m_C$ most relevant frames in $I_j^\star$
    \STATE Add the resulting Coverage-E unit to $\mathcal{U}$ and
    $I_j^\star$ to $\mathcal{O}$
\ENDFOR
\STATE Fill unassigned Coverage-E units using the highest-scoring
valid spans globally

\IF{the Dense-E or Coverage-E budget is not fully realized}
    \STATE Set $B_D$ and $B_C$ to their realized counts,
    $B_F\leftarrow B-B_D-B_C$, reset $\mathcal{U}$ and $\mathcal{O}$,
    and repeat once
\ENDIF

\STATE Sort $\mathcal{U}$ by representative timestamp
\RETURN $\mathcal{U}$
\end{algorithmic}
\end{algorithm}

\textbf{Dense-E allocation.}
ReUnit assigns a priority to each candidate span using its average relevance:
\begin{equation}
\mu(I)=\frac{1}{|I|}\sum_{t\in I}r_t.
\end{equation}
At each step, ReUnit selects the available Dense-E span with the highest priority:
\begin{equation}
I^\star=
\operatorname*{arg\,max}_{I\in\mathcal{C}_D,\; I\cap\mathcal{O}=\emptyset}
\mu(I).
\end{equation}
ReUnit then selects the four sampled frames with the highest relevance within $I^\star$ as the source frames of Dense-E. Their indices are recorded in $\mathcal{I}_D(I^\star)$ in temporal order. Together with the representative timestamp, these indices define the Dense-E specification. The selected span is then added to $\mathcal{O}$. Focal-E selects relevant moments from the full sampled sequence. Dense-E instead selects a local span first and then retains several relevant moments within that span.

\textbf{Coverage-E allocation.}
Coverage-E uses the same average-relevance score as Dense-E but additionally
encourages its spans to cover different temporal regions. ReUnit divides the
sampled timeline into $B_C$ temporal bins $\{G_j\}_{j=1}^{B_C}$. For each bin,
the available Coverage-E candidate spans are defined as
\begin{equation}
\mathcal{C}_C^{(j)}
=
\left\{
I\in\mathcal{C}_C
\mid
I\cap\mathcal{O}=\emptyset,\;
\operatorname{mid}(I)\in G_j
\right\},
\end{equation}
where $\operatorname{mid}(I)$ denotes the midpoint of span $I$. ReUnit then
prioritizes the span with the highest average relevance:
\begin{equation}
I_j^\star
=
\operatorname*{arg\,max}_{I\in\mathcal{C}_C^{(j)}}
\mu(I).
\end{equation} 
If no valid span is available in a bin, that bin is skipped. Any unfilled
Coverage-E units are subsequently assigned to the highest-priority
unoccupied spans. ReUnit then selects the nine most relevant sampled frames
within each selected span and records them in temporal order. Together with the
representative timestamp, these indices define the Coverage-E specification,
and the selected span is added to $\mathcal{O}$.

If the planned Dense-E or Coverage-E budget cannot be fully realized under
the occupancy constraint, ReUnit sets $B_D$ and $B_C$ to the realized counts
and assigns the remaining units to $B_F$. It then reruns the allocation once
from the beginning. Finally, ReUnit sorts all unit specifications by their representative timestamps $\tau_i$. Visual unitization then converts the ordered specifications into the image sequence $\widetilde{\mathcal{U}}$.

\begin{table*}[t]
\centering
\small
\caption{Comparison with representative long-video visual input construction methods on four benchmarks under four visual input budgets, using Qwen3-VL-8B-Instruct. $\Delta$ denotes the absolute improvement over Uniform Sampling at the same budget. Average is the arithmetic mean of MLVU, LongVideoBench, LVBench, and VideoMME Overall. \textbf{Bold} and \underline{underlined} values indicate the best and second-best results.}
\setlength{\tabcolsep}{2.0pt}
\renewcommand{\arraystretch}{1.08}
\setlength{\aboverulesep}{0.25ex}
\setlength{\belowrulesep}{0.25ex}
\setlength{\cmidrulekern}{3pt}

\begin{tabularx}{0.98\textwidth}{
    @{}
    >{\centering\arraybackslash}p{0.175\textwidth}
    !{\vrule width 0.5pt}
    Z{0.75} Z{0.75} Z{0.75}
    Z{0.75} Z{0.75} Z{0.75} Z{0.75}
    !{\vrule width 0.5pt}
    Z{0.58} Z{0.52}
    @{}
}
\toprule
\multirow{2}{*}{\textbf{Method}}
& \multirow{2}{*}{\textbf{MLVU}}
& \textbf{LongVideo}
& \multirow{2}{*}{\textbf{LVBench}}
& \multicolumn{4}{c!{\vrule width 0.3pt}}{\textbf{VideoMME} {\scriptsize (w/o subtitles)}}
& \multicolumn{2}{c}{\multirow{2}{*}{\textbf{Average}}} \\

\cmidrule(lr){5-8}

& &\textbf{Bench} & & \textbf{Overall} & \textbf{Short} & \textbf{Medium} & \textbf{Long} & & \\[-0.5pt]

\textit{Duration}
& \textit{3$\sim$120min}
& \textit{1$\sim$60min}
& \textit{30$\sim$120min}
& \textit{1$\sim$60min}
& \textit{1$\sim$3min}
& \textit{3$\sim$30min}
& \textit{30$\sim$60min}
& \textbf{Score}
& \textbf{$\Delta$} \\
\midrule

\multicolumn{10}{c}{\textit{\textbf{Visual Budget: 4 Units}}} \\
\midrule
\rowcolor{baselinebg}\textbf{Uniform Sampling} & 53.1 & 51.7 & 33.4 & 53.5 & 59.9 & 51.8 & 48.9 & 47.9 & -- \\
\rowcolor{baselinebg}\textbf{WFS-SB}{\scriptsize(CVPR2026)} & 63.2 & \underline{58.1} & \underline{45.5} & 58.6 & 67.1 & 56.4 & \underline{52.3} & \underline{56.4} & \underline{8.5} \\
\rowcolor{baselinebg}\textbf{QCA}{\scriptsize(ECCV2026)} & \textbf{64.3} & 55.9 & 43.1 & \underline{58.7} & \underline{67.4} & \underline{56.6} & 52.2 & 55.5 & 7.6 \\
\rowcolor{baselinebg}\textbf{VideoPanels}{\scriptsize(CVPR2026)} & 54.7 & 53.9 & 35.8 & 56.9 & 65.2 & 53.9 & 51.7 & 50.3 & 2.4 \\
\rowcolor{oursbg}\textbf{ReUnit{\scriptsize (Ours)}} & \textbf{64.3} & \textbf{58.7} & \textbf{47.2} & \textbf{60.5} & \textbf{69.8} & \textbf{58.3} & \textbf{53.4} & \textbf{57.7} & \textbf{9.8} \\

\midrule
\multicolumn{10}{c}{\textit{\textbf{Visual Budget: 8 Units}}} \\
\midrule
\rowcolor{baselinebg}\textbf{Uniform Sampling} & 55.2 & 55.9 & 37.1 & 57.0 & 65.8 & 55.1 & 50.0 & 51.3 & -- \\
\rowcolor{baselinebg}\textbf{WFS-SB}{\scriptsize(CVPR2026)} & 67.6 & \underline{61.0} & \underline{47.6} & 61.5 & \underline{71.3} & \underline{59.3} & 53.8 & \underline{59.4} & \underline{8.1} \\
\rowcolor{baselinebg}\textbf{QCA}{\scriptsize(ECCV2026)} & \underline{68.0} & 58.6 & 46.5 & \underline{62.1} & 70.3 & 59.0 & \underline{55.1} & 58.8 & 7.5 \\
\rowcolor{baselinebg}\textbf{VideoPanels}{\scriptsize(CVPR2026)} & 58.2 & 56.5 & 37.1 & 60.6 & 70.8 & 58.7 & 52.3 & 53.1 & 1.8 \\
\rowcolor{oursbg}\textbf{ReUnit{\scriptsize (Ours)}} & \textbf{68.7} & \textbf{61.7} & \textbf{51.1} & \textbf{63.3} & \textbf{73.4} & \textbf{61.0} & \textbf{55.4} & \textbf{61.2} & \textbf{9.9} \\

\midrule
\multicolumn{10}{c}{\textit{\textbf{Visual Budget: 16 Units}}} \\
\midrule
\rowcolor{baselinebg}\textbf{Uniform Sampling} & 58.8 & 58.0 & 38.0 & 60.9 & 71.0 & 58.9 & 52.8 & 53.9 & -- \\
\rowcolor{baselinebg}\textbf{WFS-SB}{\scriptsize(CVPR2026)} & \underline{70.5} & \underline{64.3} & \underline{50.7} & \underline{65.0} & \underline{75.2} & 64.2 & 55.6 & \underline{62.6} & \underline{8.7} \\
\rowcolor{baselinebg}\textbf{QCA}{\scriptsize(ECCV2026)} & 68.4 & 59.2 & 47.4 & 64.9 & 73.7 & \underline{64.3} & \textbf{56.7} & 60.0 & 6.1 \\
\rowcolor{baselinebg}\textbf{VideoPanels}{\scriptsize(CVPR2026)} & 62.3 & 58.4 & 39.7 & 63.8 & 75.1 & 62.4 & 53.8 & 56.1 & 2.2 \\
\rowcolor{oursbg}\textbf{ReUnit{\scriptsize (Ours)}} & \textbf{71.3} & \textbf{65.2} & \textbf{53.1} & \textbf{66.2} & \textbf{75.7} & \textbf{66.6} & \underline{56.4} & \textbf{64.0} & \textbf{10.1} \\

\midrule
\multicolumn{10}{c}{\textit{\textbf{Visual Budget: 32 Units}}} \\
\midrule
\rowcolor{baselinebg}\textbf{Uniform Sampling} & 63.6 & 61.0 & 39.3 & 65.5 & 77.3 & 66.1 & 56.0 & 57.4 & -- \\
\rowcolor{baselinebg}\textbf{WFS-SB}{\scriptsize(CVPR2026)} & \textbf{74.0} & \underline{65.6} & \underline{54.6} & \textbf{68.1} & \textbf{79.2} & \underline{66.9} & \underline{58.2} & \underline{65.6} & \underline{8.2} \\
\rowcolor{baselinebg}\textbf{QCA}{\scriptsize(ECCV2026)} & 71.8 & 63.4 & 52.0 & 67.8 & 76.3 & \textbf{67.0} & 57.8 & 63.8 & 6.4 \\
\rowcolor{baselinebg}\textbf{VideoPanels}{\scriptsize(CVPR2026)} & 65.1 & 61.5 & 43.4 & 66.0 & 76.1 & 65.4 & 56.3 & 59.0 & 1.6 \\
\rowcolor{oursbg}\textbf{ReUnit{\scriptsize (Ours)}} & \underline{73.5} & \textbf{66.3} & \textbf{54.7} & \textbf{68.1} & \underline{79.1} & \underline{66.9} & \textbf{58.3} & \textbf{65.7} & \textbf{8.3} \\
\bottomrule
\end{tabularx}
\label{tab:main_results}
\end{table*}


\subsection{Optional Lightweight Unit-Format Adaptation}
\label{subsec:adaptation}

By default, ReUnit requires no additional training, and the constructed image input sequence $\widetilde{\mathcal{U}}$ can be provided directly to the downstream Video-MLLM. We additionally consider a lightweight adaptation stage to evaluate how exposure to the heterogeneous unit format affects the downstream model. We use 30,000 video samples with durations of up to 180 seconds from LLaVA-Video-178K~\cite{zhang2025llavavideodata}. The original text instructions, answer annotations, and training objective are kept unchanged. Only the visual input construction is modified.

Unlike inference, where the unit type is determined according to query relevance, the adaptation stage uses a query-agnostic construction. Anchor frames are randomly sampled from the video, and unit types are assigned at random according to the prescribed allocation ratio. For Dense-E and Coverage-E, a local span is constructed around each anchor frame. The required number of source frames is then sampled uniformly from the span. The resulting unit specifications are converted into image inputs using the same unitization process described in Section~\ref{subsec:unitization}.

During adaptation, only the LLM parameters of the downstream model are updated using LoRA~\cite{hu2022lora}. All other parameters remain frozen. Unless otherwise stated, all main results use the default training-free setting.

\section{Experiments}
\label{sec:exp}

\subsection{Experimental Setup}
\label{subsec:setup}

\textbf{Benchmarks.}
We evaluate ReUnit on four long-video understanding benchmarks with diverse
video durations and reasoning requirements: MLVU~\cite{zhou2025mlvu},
LongVideoBench~\cite{wu2024longvideobench}, LVBench~\cite{wang2025lvbench},
and VideoMME~\cite{fu2025videomme}. MLVU is a multi-task benchmark covering
diverse video genres and long-video understanding capabilities, including
information retrieval, temporal reasoning, and holistic video understanding.
LongVideoBench evaluates long-context video question answering, with questions
that require locating relevant evidence from extended video contexts and
reasoning over it. LVBench focuses on extreme long-video understanding and
emphasizes information retrieval and reasoning over substantially longer video
content. VideoMME covers diverse video domains and organizes videos into short,
medium, and long duration groups, allowing us to examine performance across
different video lengths. We report multiple-choice accuracy on all benchmarks
and evaluate VideoMME without subtitles.

\begin{table}[!t]
\centering
\footnotesize
\caption{Comparison with additional long-video input construction methods under matched downstream models. Base and $+$M denote results without and with the corresponding method, and $\Delta$ denotes their difference. \textbf{Bold} and \underline{underlined} values indicate the best and second-best $\Delta$ within each budget group. $^\dagger$Q-Frame uses a multi-resolution budget ($4+8+32$).}
\label{tab:published_comparison}
\setlength{\tabcolsep}{1.2pt}
\renewcommand{\arraystretch}{1.12}
\setlength{\aboverulesep}{.25ex}
\setlength{\belowrulesep}{.25ex}

\begin{tabularx}{\columnwidth}{@{}
>{\centering\arraybackslash}p{.282\columnwidth}!{\vrule width .3pt}
>{\centering\arraybackslash}p{.07\columnwidth}!{\vrule width .3pt}
YYY!{\vrule width .3pt}YYY@{}}
\toprule
\multirow{2}{*}{\textbf{Method}}&
\multirow{2}{*}{\textbf{Units}}&
\multicolumn{3}{c!{\vrule width .3pt}}{\textbf{LongVideoBench}}&
\multicolumn{3}{c}{\textbf{VideoMME{\tiny(w/o subtitles)}}}\\
\cmidrule(lr){3-5}\cmidrule(lr){6-8}
&&\textbf{Base}&\textbf{+M}&\textbf{$\Delta$}
&\textbf{Base}&\textbf{+M}&\textbf{$\Delta$}\\
\midrule

\multicolumn{8}{c}{\textbf{\textit{Qwen2.5-VL-7B-Instruct}}}\\
\midrule
\rowcolor{baselinebg}
\textbf{GMM-EVA}{\tiny(PRCV2026)}&$\sim$16&57.4&61.2&
\dg{\underline{+3.8}}&60.9&61.1&\dg{\underline{+0.2}}\\
\rowcolor{oursbg}
\textbf{ReUnit{\tiny(Ours)}}&16&56.6&61.3&
\dg{\textbf{+4.7}}&57.9&61.5&\dg{\textbf{+3.6}}\\
\midrule
\rowcolor{baselinebg}
\textbf{MDP3}{\tiny(ICCV2025)}&32&58.1&60.0&
\dg{\underline{+1.9}}&60.8&63.8&\dg{\underline{+3.0}}\\
\rowcolor{baselinebg}
\textbf{GIFT}{\tiny(CVPR2026)}&32&59.5&61.3&
\dg{+1.8}&61.0&64.4&\dg{\textbf{+3.4}}\\
\rowcolor{oursbg}
\textbf{ReUnit{\tiny(Ours)}}&32&58.9&63.7&
\dg{\textbf{+4.8}}&61.6&63.9&\dg{+2.3}\\
\midrule

\multicolumn{8}{c}{\textbf{\textit{Qwen2-VL-7B-Instruct}}}\\
\midrule
\rowcolor{baselinebg}
\textbf{Q-Frame}{\tiny(ICCV2025)}&8$^\dagger$&53.5&58.4&
\dg{\underline{+4.9}}&53.7&58.3&\dg{\underline{+4.6}}\\
\rowcolor{oursbg}
\textbf{ReUnit{\tiny(Ours)}}&8&51.5&57.7&
\dg{\textbf{+6.2}}&52.3&57.2&\dg{\textbf{+4.9}}\\
\midrule
\rowcolor{baselinebg}
\textbf{AKS}{\tiny(CVPR2025)}&32&55.5&60.5&
\dg{+5.0}&57.6&59.9&\dg{\underline{+2.3}}\\
\rowcolor{baselinebg}
\textbf{TCS}{\tiny(ICASSP2026)}&32&55.5&60.9&
\dg{+5.4}&57.6&58.0&\dg{+0.4}\\
\rowcolor{baselinebg}
\textbf{FOCUS}{\tiny(ICLR2026)}&32&55.6&62.3&
\dg{\textbf{+6.7}}&57.6&59.7&\dg{+2.1}\\
\rowcolor{oursbg}
\textbf{ReUnit{\tiny(Ours)}}&32&55.6&61.9&
\dg{\underline{+6.3}}&57.1&60.6&\dg{\textbf{+3.5}}\\
\bottomrule
\end{tabularx}
\end{table}

\textbf{Models.}
Unless otherwise stated, the main experiments use Qwen3-VL-8B-Instruct~\cite{bai2025qwen3vltechnicalreport} as the downstream Video-MLLM. For cross-model generalization, we further test ReUnit on InternVL3.5-8B-Instruct~\cite{wang2025internvl35advancingopensourcemultimodal}, MiniCPM-o-4.5~\cite{cui2026minicpmo45realtimefullduplex}, and LLaVA-OneVision-2-8B-Instruct~\cite{an2026llavaonevision2nextgenerationperceptualintelligence}. To match the downstream models used in recent reports, we also evaluate
ReUnit with Qwen2.5-VL-7B-Instruct~\cite{bai2025qwen25vltechnicalreport}
and Qwen2-VL-7B-Instruct~\cite{wang2024qwen2vl}. All evaluations conducted in our experiments follow a unified protocol based on lmms-eval~\cite{zhang2025lmms_eval}.

\begin{table*}[t]
\centering
\small
\caption{Cross-model evaluation of ReUnit on three Video-MLLM families under visual input budgets of 8 and 32 units. Green annotations denote the absolute gain over Uniform Sampling under the same model and budget.}
\setlength{\tabcolsep}{1.3pt}
\renewcommand{\arraystretch}{1.15}
\setlength{\aboverulesep}{.25ex}
\setlength{\belowrulesep}{.25ex}
\setlength{\cmidrulekern}{0pt}
\begin{tabularx}{0.98\textwidth}{
  @{}
  >{\centering\arraybackslash}p{.16\textwidth}
  !{\vrule width 0.5pt}
  >{\centering\arraybackslash}p{.06\textwidth}
  !{\vrule width 0.5pt}
  Z{.90} Z{0.9} Z{.90}
  Z{.8} Z{.80} Z{.85} Z{.80}
  @{}
}
\toprule
\multirow{2}{*}{\textbf{Method}}
& \multirow{2}{*}{\textbf{Units}}
& \multirow{2}{*}{\textbf{MLVU}}
& \textbf{LongVideo}
& \multirow{2}{*}{\textbf{LVBench}}
& \multicolumn{4}{c}{\textbf{VideoMME}{\scriptsize (w/o subtitles)}} \\
\cmidrule{6-9}
& & &\textbf{Bench} & &
\textbf{Overall} & \textbf{Short} & \textbf{Medium} & \textbf{Long} \\
\midrule
\multicolumn{9}{c}{
  \textit{\textbf{InternVL3.5-8B-Instruct}}} \\
\midrule
\rowcolor{baselinebg}
\textbf{Uniform Sampling} & 8
& \scoreplain{61.7} & \scoreplain{53.7}
& \scoreplain{41.3} & \scoreplain{59.7}
& \scoreplain{69.2} & \scoreplain{59.3}
& \scoreplain{50.4} \\
\rowcolor{oursbg}
\textbf{ReUnit{\scriptsize (Ours)}} & 8
& \scoregain{67.0}{5.3} & \scoregain{61.6}{7.9}
& \scoregain{49.3}{8.0} & \scoregain{61.8}{2.1}
& \scoregain{72.9}{3.7} & \scoregain{60.1}{0.8}
& \scoregain{52.4}{2.0} \\
\midrule
\rowcolor{baselinebg}
\textbf{Uniform Sampling} & 32
& \scoreplain{68.3} & \scoreplain{59.9}
& \scoreplain{41.8} & \scoreplain{64.4}
& \scoreplain{76.6} & \scoreplain{63.4}
& \scoreplain{53.2} \\
\rowcolor{oursbg}
\textbf{ReUnit{\scriptsize (Ours)}} & 32
& \scoregain{70.4}{2.1} & \scoregain{65.1}{5.2}
& \scoregain{53.4}{11.6} & \scoregain{65.8}{1.4}
& \scoregain{77.6}{1.0} & \scoregain{65.7}{2.3}
& \scoregain{54.1}{0.9} \\
\midrule
\multicolumn{9}{c}{
  \textit{\textbf{MiniCPM-o-4.5}}} \\
\midrule
\rowcolor{baselinebg}
\textbf{Uniform Sampling} & 8
& \scoreplain{58.5} & \scoreplain{55.6}
& \scoreplain{38.2} & \scoreplain{58.4}
& \scoreplain{68.0} & \scoreplain{57.3}
& \scoreplain{49.9} \\
\rowcolor{oursbg}
\textbf{ReUnit{\scriptsize (Ours)}} & 8
& \scoregain{71.7}{13.2} & \scoregain{61.6}{6.0}
& \scoregain{50.5}{12.3} & \scoregain{63.8}{5.4}
& \scoregain{73.3}{5.3} & \scoregain{64.0}{6.7}
& \scoregain{54.0}{4.1} \\
\midrule
\rowcolor{baselinebg}
\textbf{Uniform Sampling} & 32
& \scoreplain{69.2} & \scoreplain{60.1}
& \scoreplain{43.5} & \scoreplain{65.9}
& \scoreplain{77.8} & \scoreplain{64.8}
& \scoreplain{55.1} \\
\rowcolor{oursbg}
\textbf{ReUnit{\scriptsize (Ours)}} & 32
& \scoregain{76.5}{7.3} & \scoregain{66.7}{6.6}
& \scoregain{54.9}{11.4} & \scoregain{67.9}{2.0}
& \scoregain{78.3}{0.5} & \scoregain{69.3}{4.5}
& \scoregain{56.0}{0.9} \\
\midrule
\multicolumn{9}{c}{
  \textit{\textbf{LLaVA-OneVision-2-8B-Instruct}}} \\
\midrule
\rowcolor{baselinebg}
\textbf{Uniform Sampling} & 8
& \scoreplain{56.9} & \scoreplain{58.6}
& \scoreplain{37.8} & \scoreplain{57.9}
& \scoreplain{67.9} & \scoreplain{56.2}
& \scoreplain{49.6} \\
\rowcolor{oursbg}
\textbf{ReUnit{\scriptsize (Ours)}} & 8
& \scoregain{72.7}{15.8} & \scoregain{65.4}{6.8}
& \scoregain{51.2}{13.4} & \scoregain{64.5}{6.6}
& \scoregain{74.7}{6.8} & \scoregain{64.2}{8.0}
& \scoregain{54.6}{5.0} \\
\midrule
\rowcolor{baselinebg}
\textbf{Uniform Sampling} & 32
& \scoreplain{67.6} & \scoreplain{63.4}
& \scoreplain{43.4} & \scoreplain{66.5}
& \scoreplain{79.2} & \scoreplain{64.7}
& \scoreplain{55.6} \\
\rowcolor{oursbg}
\textbf{ReUnit{\scriptsize (Ours)}} & 32
& \scoregain{76.6}{9.0} & \scoregain{68.1}{4.7}
& \scoregain{55.8}{12.4} & \scoregain{68.6}{2.1}
& \scoregain{80.7}{1.5} & \scoregain{68.1}{3.4}
& \scoregain{57.1}{1.5} \\
\bottomrule
\end{tabularx}
\label{tab:cross_model_generalization}
\end{table*}

\begin{figure*}[t]
    \centering
    \includegraphics[width=\textwidth]{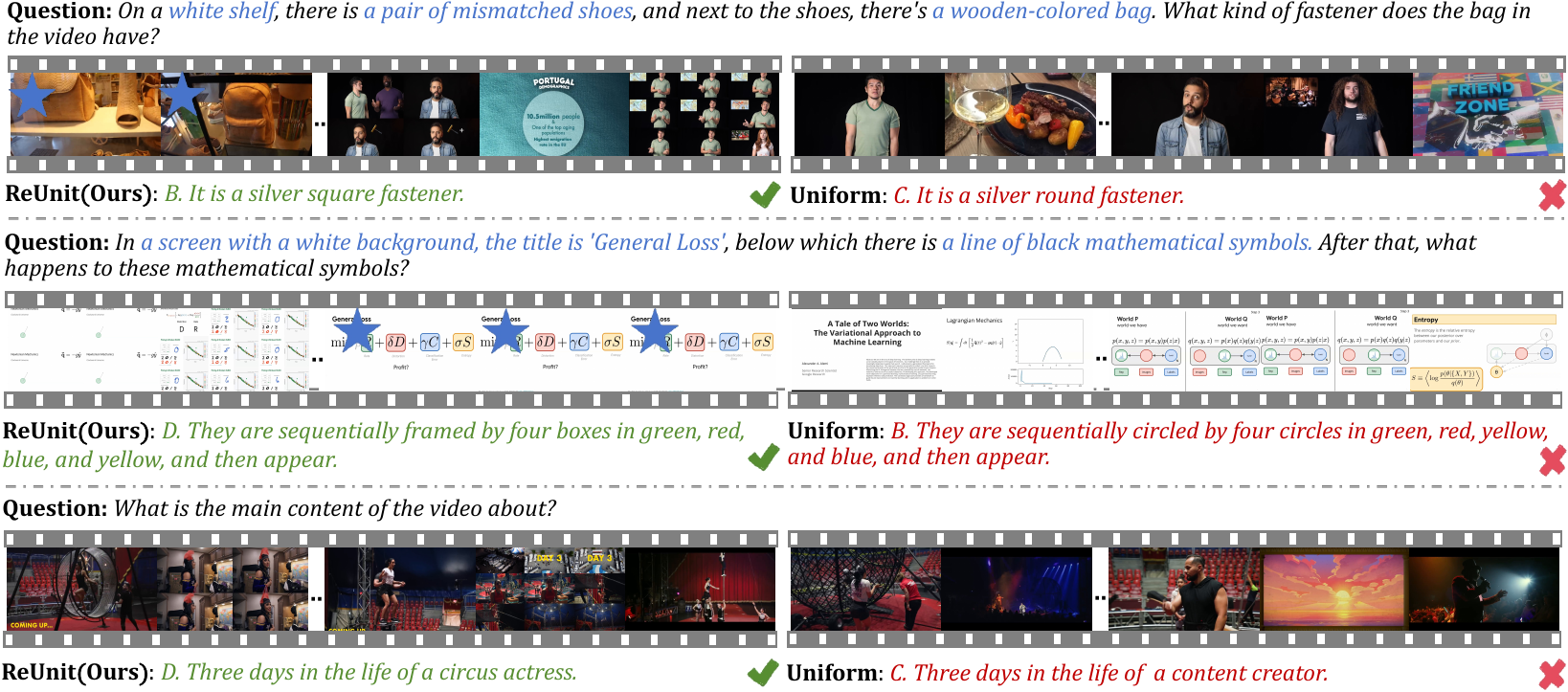}
    \caption{
    Qualitative comparison with Uniform Sampling on multiple-choice video QA
    under the same visual input budget. The three examples respectively require
    fine-grained visual detail, temporally ordered evidence across multiple
    moments, and broader contextual evidence for understanding the overall video.
    ReUnit preserves the answer-critical evidence required by these different
    queries, whereas Uniform Sampling misses or sparsely covers the relevant
    content. Blue stars indicate answer-critical visual evidence.
    }
    \label{fig:closed_end}
\end{figure*}

\textbf{Baselines.}
We compare ReUnit with Uniform Sampling and representative recent long-video input construction and frame-selection methods, including WFS-SB~\cite{chen2026wavelet}, QCA~\cite{peng2026qca}, VideoPanels~\cite{doorenbos2026videopanels}, GMM-EVA~\cite{lu2026gaussian}, MDP3~\cite{sun2025mdp3}, GIFT~\cite{ma2026gift}, Q-Frame~\cite{Zhang2025qframe}, AKS~\cite{Tang2025aks}, TCS~\cite{tan2026thinkclipsample}, and FOCUS~\cite{ziruiz2026focus}. For the main controlled comparison, we reproduce the corresponding baselines within a unified evaluation pipeline. To broaden the comparison to additional recent methods, we further include results reported in their original papers. For these supplementary comparisons, ReUnit is evaluated with the corresponding downstream model, and the budget setting of each method is explicitly reported.

\textbf{Implementation details.}
Videos are sampled at 1 FPS, and the query-relevance signal is computed using a frozen BLIP-ITM~\cite{li2022blip} scorer, with both the question and its answer options included in the textual input. We use a fixed allocation ratio
$(\alpha_F{:}\alpha_D{:}\alpha_C)=7{:}2{:}1$ and span expansion factor
$\kappa=4$ across all benchmarks and visual input budgets, without
dataset-specific tuning. Their sensitivity is analyzed in
Sec.~\ref{subsec:ablation}. We evaluate four visual input budgets $B\in\{4,8,16,32\}$. ReUnit requires no additional training by default. The optional adaptation in Sec.~\ref{subsec:adaptation} applies LoRA~\cite{hu2022lora} only to the LLM component. Benchmark evaluation is parallelized across four NVIDIA RTX A6000 GPUs, while the optional adaptation is conducted on two NVIDIA A800 GPUs. Runtime measurements are conducted separately on a single NVIDIA RTX A6000 GPU.

\subsection{Comparison with State-of-the-Art Methods}
\label{subsec:sota}

\textbf{Main comparison.}
Table~\ref{tab:main_results} presents a controlled comparison with
representative recent long-video input construction methods on four benchmarks
under four visual input budgets. All baselines are reproduced within our
unified evaluation pipeline using the same Qwen3-VL-8B-Instruct model.
ReUnit achieves the highest average score at every tested budget. Relative
to Uniform Sampling, the average score increases by $9.8$, $9.9$, $10.1$, and $8.3$ points at budgets
of 4, 8, 16, and 32 units, respectively. Compared with the strongest competing
method, ReUnit further improves the average score by $1.3$, $1.8$, and $1.4$
points at 4, 8, and 16 units. Its advantage is particularly pronounced on
LVBench, where ReUnit obtains the best result across all four budgets. As the
budget increases to 32 units, the average gap to the strongest baseline narrows
to $0.1$ points, while ReUnit still retains the highest average score. Its relative advantage is therefore larger when the available visual input budget is constrained.

\textbf{Comparison with more methods.}
Many recent methods are reported with specific Video-MLLMs and evaluation
settings, making a unified reproduction difficult.
Table~\ref{tab:published_comparison} therefore extends the comparison using
results reported in their original papers. ReUnit is evaluated with the
corresponding downstream model, and we compare the improvement over each
method's corresponding base model. For example, Q-Frame reports its
multi-resolution results using the multi-image mode of Qwen2-VL rather than
its video mode, and notes that these two modes activate different model
weights~\cite{Zhang2025qframe}. Overall, ReUnit achieves the largest gain in six of the eight
benchmark-setting comparisons and the second-largest gain in one more.

\begin{table}[t]
\centering
\footnotesize
\caption{Performance of single-granularity ReUnit variants at 16 units on Qwen3-VL-8B-Instruct. The first three rows use only Focal-E, Dense-E, or Coverage-E, while the last row uses all three granularities. \textbf{Best} results are shown in bold.}
\label{tab:single_granularity}
\setlength{\tabcolsep}{1.5pt}
\renewcommand{\arraystretch}{1.12}
\setlength{\aboverulesep}{.25ex}
\setlength{\belowrulesep}{.25ex}

\begin{tabularx}{0.95\columnwidth}{@{}
Z{.8} Z{.8} Z{1.0}!{\vrule width .3pt}Z{1.3} Z{1.06}@{}}
\toprule
\textbf{Focal-E}&\textbf{Dense-E}&\textbf{Coverage-E}&
\textbf{LongVideoBench}&\textbf{VideoMME}\\
\midrule
\textbf{$\checkmark$}&&&\textbf{65.5}&64.9\\
&\textbf{$\checkmark$}&&62.1&64.9\\
&&\textbf{$\checkmark$}&60.2&63.9\\
\rowcolor{oursbg}\textbf{$\checkmark$}&\textbf{$\checkmark$}&\textbf{$\checkmark$}&65.2&\textbf{66.2}\\
\bottomrule
\end{tabularx}
\end{table}

\subsection{Cross-Model Generalization}
\label{subsec:crossmodel}

Table~\ref{tab:cross_model_generalization} evaluates ReUnit on three
additional Video-MLLM families under visual input budgets of 8 and 32 units.
ReUnit consistently improves over Uniform Sampling across all tested models,
benchmarks, and budgets. The gains are particularly pronounced under the
8-unit budget: ReUnit improves MLVU and LVBench by $13.2$ and $12.3$ points
on MiniCPM-o-4.5, and by $15.8$ and $13.4$ points on
LLaVA-OneVision-2-8B-Instruct, respectively. On LVBench, the gains over Uniform Sampling remain substantial at 32 units:
$11.6$, $11.4$, and $12.4$ points on InternVL3.5-8B-Instruct,
MiniCPM-o-4.5, and LLaVA-OneVision-2-8B-Instruct, respectively. These consistent gains
across different model families and visual input budgets demonstrate that
ReUnit transfers effectively beyond the Qwen3-VL-8B-Instruct model used in
our main experiments.

\begin{table}[t]
\centering
\footnotesize
\caption{Leave-one-granularity-out ablation at 16 units on Qwen3-VL-8B-Instruct. The budget of the removed unit type is reallocated to the remaining granularities according to their original ratio. Uniform Sampling and the full ReUnit are included for reference.}
\label{tab:granularity_contribution}
\setlength{\tabcolsep}{2pt}
\renewcommand{\arraystretch}{1.12}
\setlength{\aboverulesep}{.25ex}
\setlength{\belowrulesep}{.25ex}

\begin{tabularx}{0.95\columnwidth}{@{}
>{\raggedright\arraybackslash}p{.42\columnwidth}
!{\vrule width .3pt}
YY@{}}
\toprule
\textbf{Configuration}&
\textbf{LongVideoBench}&
\textbf{VideoMME}\\
\midrule
\textbf{Uniform Sampling}&58.0&60.9\\
\midrule
\rowcolor{oursbg}
\textbf{ReUnit} (Our full method)&\textbf{65.2}&\textbf{66.2}\\
\hspace{.8em}w/o Focal-E&60.9&63.4\\
\hspace{.8em}w/o Dense-E&64.7&63.9\\
\hspace{.8em}w/o Coverage-E&64.8&64.1\\
\bottomrule
\end{tabularx}
\end{table}

\begin{table}[t]
\centering
\footnotesize
\caption{Granularity allocation ablation at 16 units on Qwen3-VL-8B-Instruct. 
Randomized Priority randomizes the unit allocation order, while Reverse Priority uses 
$C{\rightarrow}D{\rightarrow}F$. All variants use the same unit budget and unit construction.}
\label{tab:granularity_allocation}
\setlength{\tabcolsep}{2pt}
\renewcommand{\arraystretch}{1.12}
\setlength{\aboverulesep}{.25ex}
\setlength{\belowrulesep}{.25ex}
\begin{tabularx}{0.95\columnwidth}{@{}
>{\raggedright\arraybackslash}p{.40\columnwidth}
!{\vrule width .3pt}
YY@{}}
\toprule
\textbf{Configuration} &
\textbf{LongVideoBench} &
\textbf{VideoMME} \\
\midrule
\textbf{Uniform Sampling} &
58.0 &
60.9 \\
\midrule
\rowcolor{oursbg}
\textbf{ReUnit} (Our full method) &\textbf{65.2} &\textbf{66.2} \\
\hspace{.8em}w/ Randomized Priority &63.9 &64.5 \\
\hspace{.8em}w/ Reverse Priority &63.2 &64.5 \\
\bottomrule
\end{tabularx}
\end{table}

\subsection{Qualitative Analysis}
\label{subsec:qualitative}

Fig.~\ref{fig:closed_end} compares ReUnit with Uniform Sampling under the same
visual input budget. In the first example, answering the question requires
distinguishing the fine-grained shape of the bag fastener. ReUnit preserves
the decisive visual detail, whereas Uniform Sampling misses the relevant
moment. The second example requires tracking how the mathematical symbols are
framed across several temporally ordered moments; ReUnit retains the relevant
sequence, while Uniform Sampling captures less informative lecture content.
In the third example, identifying the overall subject of the video requires
complementary observations across different stages rather than an isolated
circus scene. Together, these examples span different evidence requirements: fine detail,
temporally ordered multi-moment evidence, and broader context. ReUnit
accommodates these requirements within the same visual input budget.

\begin{table}[t]
\centering
\footnotesize
\caption{Robustness to the vision-language scorer used for query-relevance
estimation at 16 units on Qwen3-VL-8B-Instruct. Uniform Sampling is included
for reference, and BLIP-ITM is the default scorer. \textbf{Best} results are
shown in bold.}
\label{tab:relevance_model}

\setlength{\tabcolsep}{1.5pt}
\renewcommand{\arraystretch}{1.08}
\setlength{\aboverulesep}{.2ex}
\setlength{\belowrulesep}{.2ex}

\begin{tabularx}{0.88\columnwidth}{@{}
Z{1.25}!{\vrule width .3pt}Z{1.05}Z{.82}Z{.72}@{}}
\toprule
\textbf{Scorer} &
\textbf{LongVideoBench} &
\textbf{VideoMME} &
\textbf{MLVU} \\
\midrule

\textbf{Uniform Sampling} & 58.0 & 60.9 & 58.8 \\
\midrule

BLIP-2-ITM    & 64.9 & 65.8 & \textbf{71.5} \\
CLIP-ViT-B   & 61.2 & 64.9 & 69.9 \\
SigLIP-so400m& 62.3 & 65.9 & 71.2 \\
\rowcolor{oursbg}
\textbf{BLIP-ITM}{\scriptsize (default)}
             & \textbf{65.2} & \textbf{66.2} & 71.3 \\
\bottomrule
\end{tabularx}
\end{table}

\subsection{Ablation Study and Analysis}
\label{subsec:ablation}

\textbf{Is a single presentation granularity sufficient?}
Using a single presentation granularity constrains the entire visual input to
one fixed balance between visual detail and content coverage. Under the same
16-unit budget, the single-granularity variants exhibit different performance
patterns across LongVideoBench and VideoMME. No single presentation
granularity performs best on both benchmarks. The Dense-E-only variant is particularly informative because it retains
query-guided selection while using a single packed multi-frame representation
throughout the visual input. Its lower performance than the full ReUnit on
both benchmarks suggests that a query-aware selector still benefits from
allowing presentation granularity to vary across retained content. The full
ReUnit remains competitive on LongVideoBench and achieves the best result on
VideoMME (Table~\ref{tab:single_granularity}), supporting the use of multiple
presentation granularities.

\textbf{Complementarity of the three granularities.}
If the three granularities were interchangeable, reallocating the budget of a
removed unit type to the remaining types would be expected to preserve
performance. Removing Focal-E causes the largest degradation, consistent with
the importance of fine-grained visual evidence for highly query-relevant
moments (Table~\ref{tab:granularity_contribution}). Removing Dense-E or
Coverage-E also lowers performance, consistent with reduced capacity to
represent multiple moments within local spans or broader temporal context,
respectively. The three presentation granularities are therefore complementary rather than
interchangeable.

\textbf{Does allocation priority matter?}
Using multiple presentation granularities raises a further question: which
content should receive each granularity? ReUnit follows the
$F{\rightarrow}D{\rightarrow}C$ priority so that the most query-relevant
moments receive finer presentation before coarser granularities are allocated
to the remaining temporal regions. With the unit budget and unit construction
fixed, both randomizing and reversing this priority reduce performance on
LongVideoBench and VideoMME (Table~\ref{tab:granularity_allocation}).
Allocation priority therefore affects performance when multiple unit types are
combined.

Taken together, these ablations isolate different parts of the design: single-granularity
variants show distinct performance patterns, leave-one-out variants show
non-interchangeable contributions, and priority variants show that the
allocation order also matters. They support combining multiple presentation granularities with structured
allocation in ReUnit.

\textbf{Robustness to the relevance scorer.}
ReUnit uses the query-relevance signal to guide visual-unit allocation.
Table~\ref{tab:relevance_model} evaluates four vision-language scorers for
estimating this signal. Replacing the default BLIP-ITM~\cite{li2022blip}
with BLIP-2-ITM~\cite{li2023blip2}, SigLIP~\cite{zhai2023siglip}, or
CLIP~\cite{radford2021clip} consistently improves over Uniform Sampling on
all three benchmarks. Even with CLIP, which gives the lowest ReUnit results
among the tested scorers, performance remains $3.2$, $4.0$, and $11.1$
points above Uniform Sampling on LongVideoBench, VideoMME, and MLVU,
respectively. The absolute performance nevertheless varies with the relevance
model: BLIP-ITM performs best on LongVideoBench and VideoMME, whereas
BLIP-2-ITM obtains the highest result on MLVU. These results indicate that ReUnit is not tied to the default relevance scorer, although its absolute performance still depends on the relevance
signal used for allocation.

\begin{table}[t]
\centering
\footnotesize
\caption{Sensitivity to the allocation ratio $(\alpha_F{:}\alpha_D{:}\alpha_C)$ at 16 units on Qwen3-VL-8B-Instruct. The default ratio $7{:}2{:}1$ is highlighted. \textbf{Best} results are shown in bold.}
\label{tab:budget_ratio}
\setlength{\tabcolsep}{2pt}
\renewcommand{\arraystretch}{1.12}
\setlength{\aboverulesep}{.25ex}
\setlength{\belowrulesep}{.25ex}

\begin{tabularx}{.95\columnwidth}{@{}
Z{1.0}!{\vrule width .3pt}Z{1.15}Z{1.05}Z{.8}@{}}
\toprule
\textbf{Ratio}&
\textbf{LongVideoBench}&
\textbf{VideoMME}&
\textbf{MLVU}\\
\midrule
\rowcolor{oursbg}
$\mathbf{7{:}2{:}1}$&\textbf{65.2}&66.2&71.3\\
$8{:}1{:}1$&64.5&65.9&71.5\\
$6{:}2{:}2$&65.1&66.2&70.6\\
$6{:}3{:}1$&64.7&\textbf{66.3}&\textbf{71.8}\\
\bottomrule
\end{tabularx}
\end{table}

\begin{table}[t]
\centering
\footnotesize
\caption{Sensitivity to the span expansion factor $\kappa$ at 16 units on Qwen3-VL-8B-Instruct. The default setting $\kappa=4$ is highlighted. \textbf{Best} results are shown in bold.}
\label{tab:span_expansion}
\setlength{\tabcolsep}{2pt}
\renewcommand{\arraystretch}{1.12}
\setlength{\aboverulesep}{.25ex}
\setlength{\belowrulesep}{.25ex}

\begin{tabularx}{.95\columnwidth}{@{}
Z{.8}!{\vrule width .3pt}Z{1.15}Z{1.05}Z{1.0}@{}}
\toprule
\textbf{$\kappa$}&
\textbf{LongVideoBench}&
\textbf{VideoMME}&
\textbf{MLVU}\\
\midrule
$3$&64.6&65.7&\textbf{71.3}\\
\rowcolor{oursbg}
$\mathbf{4}$&65.2&\textbf{66.2}&\textbf{71.3}\\
$5$&\textbf{65.3}&66.1&70.4\\
\bottomrule
\end{tabularx}
\end{table}

\textbf{Sensitivity to hyperparameters.}
We use the same allocation ratio and span expansion factor across all
benchmarks without dataset-specific tuning. As shown in Tables~\ref{tab:budget_ratio} 
and~\ref{tab:span_expansion},
varying the allocation ratio changes accuracy by at most $1.2$ points across
the three benchmarks. The default $7{:}2{:}1$ ratio performs best on LongVideoBench, whereas
$6{:}3{:}1$ achieves slightly higher results on VideoMME and MLVU, indicating
that different benchmarks may favor different relative allocations without
substantially affecting overall performance. Varying $\kappa$ from $3$ to $5$
results in a maximum variation of $0.9$ points: $\kappa=5$ is marginally
better on LongVideoBench, while the default $\kappa=4$ performs best on
VideoMME and ties for the best result on MLVU. No alternative setting dominates all three benchmarks; the default
configuration is therefore a stable operating point without dataset-specific
hyperparameter tuning.

\begin{table}[t]
\centering
\footnotesize
\caption{Comparison between the default training-free ReUnit and the optional LoRA-adapted variant on Qwen3-VL-8B-Instruct at a budget of 32 units. \textbf{Best} results are shown in bold.}
\label{tab:finetuning}
\setlength{\tabcolsep}{5pt}
\renewcommand{\arraystretch}{1.12}

\begin{tabularx}{0.98\columnwidth}{@{}
>{\raggedright\arraybackslash}p{.28\columnwidth}
!{\vrule width .3pt}
cccc
@{}}
\toprule
\multirow{2}{*}{\textbf{Method}}&
\multirow{2}{*}{\textbf{MLVU}}&
\textbf{LongVideo}&
\multirow{2}{*}{\textbf{VideoMME}}&
\multirow{2}{*}{\textbf{LVBench}}\\[-1pt]
&&\textbf{Bench}&&\\
\midrule
\rowcolor{baselinebg}
\textbf{Uniform Sampling}&63.6&61.0&65.5&39.3\\
\midrule

\rowcolor{oursbg!55!white}
\textbf{ReUnit{\scriptsize (default)}}&73.5&\textbf{66.3}&\textbf{68.1}&54.7\\
\rowcolor{oursbg}
\textbf{ReUnit{\scriptsize(adapted)}}&\textbf{73.9}&\textbf{66.3}&67.8&\textbf{54.9}\\
\bottomrule
\end{tabularx}
\end{table}

\textbf{Optional unit-format adaptation.}
Introducing the lightweight adaptation of Sec.~\ref{subsec:adaptation}
changes performance only slightly across the four benchmarks, with differences
ranging from $-0.3$ to $+0.4$ points relative to the default ReUnit
(Table~\ref{tab:finetuning}). This limited change suggests that
Qwen3-VL-8B-Instruct can make effective use of the heterogeneous unit format
without additional ReUnit-specific adaptation. Thus, the gains of the default
ReUnit are obtained without extra training of the downstream model.

\subsection{Practical Analysis}
\textbf{Efficiency.}
Table~\ref{tab:efficiency} reports the runtime on LVBench, where each video
contains 4,046 sampled frames on average, corresponding to approximately
67.4 minutes under the 1-FPS sampling protocol. For ReUnit, query-relevance
extraction accounts for most of the processing time, while the allocation
itself requires only $0.02$ s. Compared with QCA, ReUnit requires more vision
preprocessing because Dense-E and Coverage-E incorporate multiple source
frames into each visual unit. Representing more video content within the same
final input budget therefore requires additional frames to be decoded and
processed during input construction. The final visual input remains fixed at
32 units, and the downstream MLLM inference time remains unchanged at
$3.2$ s. Overall, ReUnit requires about $31.9$ s per video--question pair, compared
with $28.5$ s for QCA and $49.9$ s for WFS-SB. Thus, the additional cost of
constructing multi-frame visual units remains modest relative to the
query-relevance extraction stage, which dominates the total runtime.

\begin{table}[t]
\centering
\footnotesize
\caption{Per video--question pair runtime breakdown on LVBench at a budget of 32 units, reported in seconds. Videos contain 4{,}046 sampled frames on average, corresponding to approximately 67.4 minutes of video under the 1-FPS sampling protocol. ITM extraction uses a batch size of 256, and all runtimes are measured on a single NVIDIA RTX A6000 GPU.}
\label{tab:efficiency}
\setlength{\tabcolsep}{2pt}
\renewcommand{\arraystretch}{1.12}
\setlength{\aboverulesep}{.25ex}
\setlength{\belowrulesep}{.25ex}

\begin{tabularx}{.95\columnwidth}{@{}
Z{.9}!{\vrule width .3pt}Z{1.5}Z{.85}Z{1.0}Z{.75}@{}}
\toprule
\multirow{2}{*}{\textbf{Method}}&
\textbf{ITM Signal}&
\multirow{2}{*}{\textbf{Algorithm}}&
\textbf{Vision}&
\textbf{MLLM}\\
&\textbf{Extraction}&&\textbf{Preprocess}&\textbf{Inference}\\
\midrule
\rowcolor{baselinebg}
\textbf{WFS-SB}&37.3&5.2&4.2&3.2\\
\rowcolor{baselinebg}
\textbf{QCA}&20.7&0.1&4.5&3.2\\
\rowcolor{oursbg}
\textbf{ReUnit}&20.7&0.02&8.0&3.2\\
\bottomrule
\end{tabularx}
\end{table}

\begin{figure}[t]
    \centering
    \includegraphics[width=0.95\linewidth]{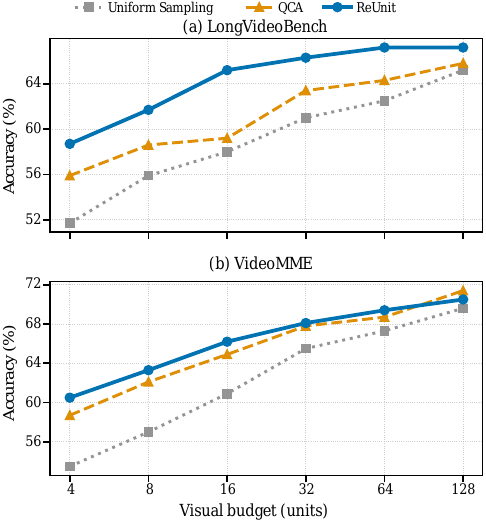}
    \caption{Accuracy of Uniform Sampling, QCA, and ReUnit under visual input budgets from 4 to 128 units on LongVideoBench and VideoMME.}
    \label{fig:limitation}
\end{figure}

\textbf{When is ReUnit most beneficial?}
The budget sweep shows the clearest relative advantage when the visual input
budget is small compared with the amount of video content to be represented. As the budget increases from
4 to 128 units, its margin over Uniform Sampling tends to narrow on both
LongVideoBench and VideoMME (Fig.~\ref{fig:limitation}). On VideoMME, QCA
catches up as the budget increases and surpasses ReUnit at 128 units, whereas
ReUnit remains ahead across all tested budgets on the longer LongVideoBench.
As more visual inputs become available, the representation constraint becomes
less severe, which is consistent with the narrowing margin. ReUnit's largest
relative gains therefore occur in the budget-constrained regime.

\section{Conclusion}
ReUnit treats long-video visual input construction under a limited visual
input budget as the joint problem of deciding which content to retain and how
it should be presented. It realizes multi-granularity visual unitization
through heterogeneous visual units that each consume one unit of the visual
input budget, with query relevance guiding their allocation. Experiments
across four benchmarks and multiple Video-MLLMs show consistent gains,
particularly under constrained visual budgets, and effective transfer across
the tested model families. The results also identify its operating regime:
ReUnit is most beneficial when the available visual budget is small relative
to the amount of video content to be represented, while its absolute
performance remains influenced by the relevance signal.

\bibliographystyle{IEEEtran}
\bibliography{references}

\end{document}